\documentclass{svproc}
\usepackage{amsfonts}
\usepackage{amsmath}
\usepackage{bm}
\usepackage{graphicx}
\usepackage{url}
\usepackage{caption}
\usepackage{subcaption}

\begin{document}

\title{Closed Form Time Derivatives of the Equations of Motion of Rigid Body
Systems}
\author{Andreas M\"uller\inst{1} \and Shivesh Kumar\inst{2}}

\titlerunning{Closed Form Time Derivatives of Equations of Motion} 
\institute{Johannes Kepler University, Linz, Austria\\
\email{a.mueller@jku.at}
\and
Robotics Innovation Center, DKFI GmbH, Bremen, Germany\\
\email{shivesh.kumar@dfki.de}}

\maketitle

\begin{abstract}
Derivatives of equations of motion(EOM) describing the dynamics of rigid
body systems are becoming increasingly relevant for the robotics community
and find many applications in design and control of robotic systems.
Controlling robots, and multibody systems comprising elastic components in
particular, not only requires smooth trajectories but also the time
derivatives of the control forces/torques, hence of the EOM. This paper
presents the time derivatives of the EOM in closed form up to second-order
as an alternative formulation to the existing recursive algorithms for this
purpose, which provides a direct insight into the structure of the
derivatives. The Lie group formulation for rigid body systems is used giving
rise to very compact and easily parameterized equations.
\end{abstract}


\section{Introduction}

Rigid body dynamics algorithms for evaluating the equations of motion (EOM)
and their derivatives find numerous applications in the design optimization
and control of modern robotic systems. The equations of motion can be
differentiated with respect to state variables, control output (generalized
forces), time and physical parameters of the robot (see~\cite{Ott2013} for
an overview). These derivatives can be computed with several methods: 1)
approximation by finite differences, 2) automatic differentiation \cite%
{AutDiff}, i.e. by applying the chain rule formula in an automatic way
knowing the derivatives of basic functions (cos, sin or exp), 3) closed form
derivatives of the EOM and 4) recursive and analytical formulations
exploiting the structure of the closed form motion equations. While the
first two methods are generic and numerical in nature, the latter two are
analytical in nature and exploit the structure of the EOM Analytical and
recursive partial derivatives of the EOM of rigid body systems with respect
to state variables and generalized forces have been reported in the
literature~\cite{Park2005,Mansard2018}. These are useful in optimal control
of legged robots (e.g. differential dynamic programming~\cite{Mansard2019})
and their computational design and optimization~\cite{Yamane2017}. Time
derivatives of EOM are required for the model-based control and motion
planning of robots with higher-order continuity \cite{ReiterTII2018}, since
for highly dynamic applications not only the actuation forces but also their
derivatives must be bounded in order to ensure feasibility. Flatness-based
control of robots with flexible joints, and of robots equipped with series
elastic actuators (SEA) or variable stiffness actuators (VSA), necessitate
the first and second time derivatives of the EOM of the robot \cite%
{deLuca1998,GattringerMUBO2014,PalliMelchiorriDeLuca2008}. Therefore,
recursive $O\left( n\right) $-algorithms for the evaluation of the time
derivatives were developed \cite%
{BuondonnaDeLuca2015,BuondonnaDeLuca2016,Guarino2006,Guarino2009,ICRA2017,RAL2020}
extending existing $O\left( n\right) $-formulations for the evaluation of
EOM. While $O\left( n\right) $-formulations are deemed computationally
advantageous when dealing with large systems, formulating and evaluating the
EOM in closed form remains an efficient alternative for many robotic systems
and provides insights into the structure of the problem. Yet, such closed
form formulations were not reported in the literature, with the exception of~%
\cite{Jain2011} where first-order time derivatives of the EOM were presented
within the so-called spatial operator framework. A relatively recent
research topic, where higher-order derivatives of the EOM are required, is
the dynamic balancing of articulated mechanisms \cite%
{VanderWijk2013,WuGosseling2004,DeJongARK2018}. In \cite{BalancingMMT2021},
the time derivatives of the spatial momentum were used to derive global
balancing conditions. Recently, we also proposed nth order time derivatives of EOM in both recursive and closed forms~\cite{KumarMueller2021}. 

In this paper, the first and second time derivatives of the EOM are
presented. 
For the subsequent treatment, the EOM are written in the form suitable for
solving the inverse dynamics problem%
\begin{equation}
\mathbf{Q}=\mathbf{M}\left( \mathbf{q}\right) \ddot{\mathbf{q}}+\mathbf{C}%
\left( \dot{\mathbf{q}},\mathbf{q}\right) \dot{\mathbf{q}}+\mathbf{Q}_{%
\mathrm{grav}}\left( \mathbf{q}\right) +\mathbf{Q}_{\mathrm{ext}}\left( 
\mathbf{q},t\right)  \label{EOM1}
\end{equation}%
where the vector of generalized coordinates $\mathbf{q}=\left( q_{1},\ldots
,q_{n}\right) ^{T}$ comprises the $n$ joint variables, $\mathbf{M}$ and $%
\mathbf{C}$ is the generalized mass and Coriolis matrix, respectively, and $%
\mathbf{Q}_{\mathrm{grav}}$ represents generalized gravity forces. The
generalized forces due to external loads (e.g. interaction/contact forces
and torques) are summarized in $\mathbf{Q}_{\mathrm{ext}}$. Finally, $%
\mathbf{Q}$ are the generalized forces (drive forces/torques) required for a
prescribed motion $\mathbf{q}\left( t\right) $.

In the following, the derivatives of (\ref{EOM1}) are derived using the Lie
group formalism reported in \cite{MUBOScrews2}, which is equivalent to those
presented in \cite{LynchPark2017} and \cite{Murray}. A salient feature of
the Lie group formulation is that it admits model description in terms of
readily available data without compromising computational efficiency (of
closed form expressions as well as $O\left( n\right) $ algorithms). As a
side contribution, we also prove the structural properties of EOM from a
closed form perspective. For sake of simplicity, and without loss of
generality a single serial kinematic chain, comprising 1-DOF joints, mounted
at the ground is considered. The generalizations to systems with arbitrary
tree-topologies is straightforward, but will not be considered here in order
to simplify notation and make the paper easily accessible.

\paragraph{Organization:}

Section~\ref{sec_EOM_closed_form} presents the equations of motion of serial
kinematic chain in closed form using the body--fixed representation of the
twists. Section~\ref{sec_first_order} and Section~\ref{sec_second_order}
presents the first- and second-order time derivatives of the equations of
motion in closed form respectively. Section~\ref{sec_structural_properties}
proves the structural properties of the EOM using the closed form
formulations. Section~\ref{sec_results} presents the application of the
proposed derivatives in evaluating second-order inverse dynamics of two
exemplary robot manipulators and a discussion on its computational
performance. Section~\ref{sec_conclusion} concludes the paper.

\section{Equations of Motion in Closed Form}

\label{sec_EOM_closed_form}

\subsection{Kinematics in terms of joint screws}

In the following, the notation and formulation of the EOM are adopted from 
\cite{MUBOScrews2}. The configuration (pose, posture) of body $i=1,\ldots ,n$
is denoted $\mathbf{C}_{i}\in SE\left( 3\right) $, which describes the frame
transformation from a body-fixed frame (arbitrarily located at the body) to
the inertial frame. Bodies and joints are numbered increasing order starting
from the ground, so that joint $i$ links body $i$ to its predecessor $i-1$,
while the ground is indexed with 0, and by convention $\mathbf{C}_{0}=%
\mathbf{I}$. The relative configuration of body $j$ with respect to body $i$
is then 
\begin{eqnarray}
\mathbf{C}_{i,j} &=&\mathbf{C}_{i}^{-1}\mathbf{C}_{j}  \notag \\
&=&\left( 
\begin{array}{cc}
\mathbf{R}_{ij} & {^{i}}\mathbf{r}_{i,j} \\ 
\mathbf{0} & 1%
\end{array}%
\right)   \label{Cij}
\end{eqnarray}%
where $\mathbf{R}_{ij}\in SO\left( 3\right) $ is the rotation matrix
transforming coordinates expressed in the reference frame on body $j$ to
their expression in the reference frame on body $i$, and ${^{i}}\mathbf{r}%
_{i,j}\in {\mathbb{R}}^{3}$ is the position vector from the origin of frame $%
i$ to the origin of frame $j$ expressed in frame $i$. Without loss of
generality, all joints are assumed to have 1-DOF. Denote with $q_{i}$ the
joint variable (rotation angle, translation) of joint $i$. The configuration
of body $i$ is determined by the product of exponential (POE) as%
\begin{equation}
\mathbf{C}_{i}\left( \mathbf{q}\right) =\mathbf{B}_{1}\exp ({^{1}\mathbf{X}}%
_{1}q_{1})\mathbf{B}_{2}\exp ({^{2}\mathbf{X}}_{2}q_{2})\cdot \ldots \cdot 
\mathbf{B}_{i}\exp ({^{i}\mathbf{X}}_{i}q_{i})  \label{Ci}
\end{equation}%
where ${\mathbf{B}}_{i}:=\mathbf{C}_{i-1,i}\left( \mathbf{0}\right) \in
SE\left( 3\right) $ is the configuration of body $i$ relative to its
predecessor $i-1$, in the reference configuration $\mathbf{q}=\mathbf{0}$,
and ${^{i}\mathbf{X}}_{i}$ is the screw coordinate vector associated to
joint $i$ represented in the body-frame of body $i$ \cite{MUBOScrews1}. The
vectors ${^{i}\mathbf{X}}_{i}$ are constant due to the body-fixed
representation. For 1-DOF lower-pair joints they are given as%
\begin{equation}
{^{i}\mathbf{X}}_{i}=\left( 
\begin{array}{c}
{^{i}}\mathbf{e}_{i} \\ 
\ {^{i}}\mathbf{x}_{i}\times {^{i}}\mathbf{e}_{i}+{^{i}}\mathbf{e}_{i}h_{i}%
\end{array}%
\right) =:\left( 
\begin{array}{c}
{^{i}}\bm{\xi}_{i} \\ 
{^{i}}\bm{\eta}_{i}%
\end{array}%
\right)   \label{Xi}
\end{equation}%
where ${^{i}}\mathbf{e}_{i}\in {\mathbb{R}}^{3}$ is a unit vector along the
joint axis, ${^{i}}\mathbf{x}_{i}\in {\mathbb{R}}^{3}$ is the vector to a
point on this axis, and $h_{i}\in {\mathbb{R}}$ is the pitch of the joint.
In particular, for a revolute and prismatic joint, the screw coordinate
vector is, respectively%
\begin{equation}
\mathrm{Revolute:\ }{^{i}\mathbf{X}}_{i}=\left( 
\begin{array}{c}
{^{i}}\mathbf{e}_{i} \\ 
\ {^{i}}\mathbf{x}_{i}\times {^{i}}\mathbf{e}_{i}%
\end{array}%
\right) ,\ \ \ \mathrm{Prismatic:\ }{^{i}\mathbf{X}}_{i}=\left( 
\begin{array}{c}
\mathbf{0} \\ 
{^{i}}\mathbf{e}_{i}%
\end{array}%
\right) .
\end{equation}%
In terms of these screw coordinates ${\mathbf{X}}$, the exponential map
attains the explicit form \cite{LynchPark2017,CND2017}%
\begin{eqnarray}
\exp (\varphi \mathbf{X}) &=&\left( 
\begin{array}{cc}
\exp (\varphi \widetilde{\mathbf{e}}) & \ \ \ (\mathbf{I}-\exp (\varphi 
\widetilde{\mathbf{e}}))\mathbf{x}+\varphi h\mathbf{e} \\ 
\mathbf{0} & 1%
\end{array}%
\right) ,\ \mathrm{for\ \ }h\neq \infty  \\
&=&\left( 
\begin{array}{cc}
\mathbf{I} & \ \varphi \mathbf{e} \\ 
\mathbf{0} & 1%
\end{array}%
\right) ,\ \mathrm{for\ \ }h=\infty   \notag
\end{eqnarray}%
where the rotation matrix is determined by the Euler-Rodrigues formula%
\begin{equation}
\exp (\varphi \widetilde{\mathbf{e}})=\mathbf{I}+\sin \varphi \widetilde{%
\mathbf{e}}+\left( 1-\cos \varphi \right) \,\widetilde{\mathbf{e}}^{2}.
\label{SO3exp}
\end{equation}

Denote with $\mathbf{V}_{i}^{\text{\textsf{b}}}=(\bm{\omega}_{i}^{\text{%
\textsf{b}}},\mathbf{v}_{i}^{\text{\textsf{b}}})^{T}$ the twist of body $i$
represented in the body-fixed frame. The superscript \textsf{b} is used to
indicate the body-fixed representation \cite{MUBOScrews2,Murray}. An
alternative representation is so-called spatial representation of twist,
which would be indicated by a superscript \textsf{s}. A recursive $O\left(
n\right) $-algorithm for evaluating the EOM and their higher-order
derivatives using the spatial representation was reported in \cite{RAL2020}.
In \cite{Kumar2019} the EOM were presented in closed form in terms of the
spatial representation of twists. The potential advantage of using the
spatial representation to express the EOM in closed form remains to be
explored, however. In this paper the (classical) body-fixed representation
is used.

The individual twists of all bodies are summarized in the vector $\mathsf{V}%
\in {\mathbb{R}}^{6n}$, which is referred to as the system twist in
body-fixed representation. It is determined as%
\begin{equation}
\mathsf{V}=\mathsf{J}\dot{\mathbf{q}}  \label{Vsys}
\end{equation}%
with the system Jacobian $\mathsf{J}\left( \mathbf{q}\right) $. The latter
admits the factorization%
\begin{equation}
\mathsf{J}=\mathsf{AX}  \label{JbSys}
\end{equation}%
in terms of the block-triangular and block-diagonal matrices 
\begin{equation}
\mathsf{A}\left( \mathbf{q}\right) =\left( 
\begin{array}{ccccc}
\mathbf{I} & \mathbf{0} & \mathbf{0} &  & \mathbf{0} \\ 
\mathbf{Ad}_{\mathbf{C}_{2,1}} & \mathbf{I} & \mathbf{0} & \cdots & \mathbf{0%
} \\ 
\mathbf{Ad}_{\mathbf{C}_{3,1}} & \mathbf{Ad}_{\mathbf{C}_{3,2}} & \mathbf{I}
&  & \mathbf{0} \\ 
\vdots & \vdots & \ddots & \ddots &  \\ 
\mathbf{Ad}_{\mathbf{C}_{n,1}} & \mathbf{Ad}_{\mathbf{C}_{n,2}} & \cdots & 
\mathbf{Ad}_{\mathbf{C}_{n,n-1}} & \mathbf{I}%
\end{array}%
\right) ,\mathsf{X}=\left( 
\begin{array}{ccccc}
{^{1}\mathbf{X}}_{1} & \mathbf{0} & \mathbf{0} &  & \mathbf{0} \\ 
\mathbf{0} & {^{2}\mathbf{X}}_{2} & \mathbf{0} & \cdots & \mathbf{0} \\ 
\mathbf{0} & \mathbf{0} & {^{3}\mathbf{X}}_{3} &  & \mathbf{0} \\ 
\vdots & \vdots & \ddots & \ddots &  \\ 
\mathbf{0} & \mathbf{0} & \cdots & \mathbf{0} & {^{n}}\mathbf{X}_{n}%
\end{array}%
\right) .  \label{Ab}
\end{equation}%
The matrix $\mathbf{Ad}_{\mathbf{C}_{i,j}}$ transforms screw coordinates
represented in the reference frame at body $j$ to those represented in the
frame on body $i$ \cite{LynchPark2017,Murray,Selig}. With the relative
configuration (\ref{Cij}) this matrix is%
\begin{equation}
\mathbf{Ad}_{\mathbf{C}_{i,j}}=\left( 
\begin{array}{cc}
\mathbf{R}_{ij} & \mathbf{0} \\ 
{^{i}}\tilde{\mathbf{r}}_{i,j}\mathbf{R}_{ij} & \mathbf{R}_{ij}%
\end{array}%
\right)  \label{AdCij}
\end{equation}%
where $\tilde{\mathbf{x}}\in so\left( 3\right) $ is the skew-symmetric
matrix associated to vector $\mathbf{x}\in {\mathbb{R}}^{3}$ so that $\tilde{%
\mathbf{x}}\mathbf{y}=\mathbf{x}\times \mathbf{y}$. A central relation for
deriving the EOM in closed form is the following expression for the time
derivative of the matrix $\mathsf{A}$ and thus of the system Jacobian \cite%
{MUBOScrews2}%
\begin{equation}
\dot{\mathsf{J}}\left( \mathbf{q},\dot{\mathbf{q}}\right) =-\mathsf{A}\left( 
\mathbf{q}\right) \mathsf{a}\left( \dot{\mathbf{q}}\right) \mathsf{J}\left( 
\mathbf{q}\right)  \label{Jdot}
\end{equation}%
where 
\begin{equation}
\mathsf{a}\left( \dot{\mathbf{q}}\right) :=\mathrm{diag}~(\dot{q}_{1}\mathbf{%
ad}_{{{^{1}}\mathbf{X}_{1}}},\ldots ,\dot{q}_{n}\mathbf{ad}_{{{^{n}}\mathbf{X%
}_{n}}}).  \label{a}
\end{equation}%
Therein, the matrix $\mathbf{ad}_{{{^{i}}\mathbf{X}_{i}}}$ is given in terms
of the joint screw coordinate vector (\ref{Xi}) as%
\begin{equation}
\mathbf{ad}_{{^{i}\mathbf{X}}_{i}}=\left( 
\begin{array}{cc}
{^{i}}\tilde{\bm{\xi}}_{i} & \ \mathbf{0} \\ 
{^{i}}\tilde{\bm{\eta}}_{i} & {^{i}}\tilde{\bm{\xi}}_{i}%
\end{array}%
\right) .  \label{adSE3}
\end{equation}%
This gives rise to the closed form expressions for the system acceleration%
\begin{equation}
\dot{\mathsf{V}}=\mathsf{J}\ddot{\mathbf{q}}-\mathsf{AaJ}\dot{\mathbf{q}}=%
\mathsf{J}\ddot{\mathbf{q}}-\mathsf{AaV}.  \label{VbdotMat}
\end{equation}%
For calculating the derivatives, the time derivative of matrix $\mathsf{A}$
will be needed. To this end, the expression%
\begin{equation}
\mathsf{A}=\left( \mathbf{I}-\mathsf{D}\right) ^{-1}  \label{A}
\end{equation}%
is used \cite{MUBOScrews2}, with the matrix%
\begin{equation}
\mathsf{D}\left( \mathbf{q}\right) =\left( 
\begin{array}{ccccc}
\mathbf{0} & \mathbf{0} & \mathbf{0} &  & \mathbf{0} \\ 
\mathbf{Ad}_{\mathbf{C}_{2,1}} & \mathbf{0} & \mathbf{0} & \cdots & \mathbf{0%
} \\ 
\mathbf{0} & \mathbf{Ad}_{\mathbf{C}_{3,2}} & \mathbf{0} &  & \mathbf{0} \\ 
\vdots & \vdots & \ddots & \ddots &  \\ 
\mathbf{0} & \mathbf{0} & \cdots & \mathbf{Ad}_{\mathbf{C}_{n,n-1}} & 
\mathbf{0}%
\end{array}%
\right) .  \label{D}
\end{equation}%
With (\ref{A}), the derivative of $\mathsf{A}$ is then%
\begin{equation}
\dot{\mathsf{A}}=\left( \mathbf{I}-\mathsf{D}\right) ^{-1}\dot{\mathsf{D}}%
\left( \mathbf{I}-\mathsf{D}\right) ^{-1}=\mathsf{A}\dot{\mathsf{D}}\mathsf{A%
}.
\end{equation}%
Using the relation $\dot{\mathbf{Ad}}_{\mathbf{C}_{i,j}}=-\dot{q}_{i}\mathbf{%
ad}_{{^{i}\mathbf{X}}_{i}}\mathbf{Ad}_{\mathbf{C}_{i,-1}}$ \cite{MUBOScrews2}%
, the derivative of $\mathsf{D}$ attains the closed form $\dot{\mathsf{D}}=-%
\mathsf{aD}$. Finally, with $\mathsf{D}=\mathbf{I}-\mathsf{A}^{-1}$, it
follows%
\begin{equation}
\dot{\mathsf{A}}\left( \mathbf{q},\dot{\mathbf{q}}\right) =\mathsf{A}\left( 
\mathbf{q}\right) \mathsf{a}-\mathsf{A}\left( \mathbf{q}\right) \mathsf{a}%
\left( \dot{\mathbf{q}}\right) \mathsf{A}\left( \mathbf{q}\right) .
\label{Adot}
\end{equation}%
Clearly, the derivative (\ref{Jdot}) of the system Jacobian is recovered as $%
\dot{\mathsf{J}}=\dot{\mathsf{A}}\mathsf{X}$ noting that $\mathsf{aX}\equiv 
\mathbf{0}$.

\subsection{Equations of Motion}

\label{sec_eom_formulae} The generalized mass and Coriolis matrix in the EOM
(\ref{EOM1}) of a simple kinematic chain mounted at the ground are found via
Jourdain's principle of virtual power as (or likewise as the Lagrange
equations) \cite{MUBOScrews2}%
\begin{equation}
\mathbf{M}\left( \mathbf{q}\right) =\mathsf{J}^{T}\mathsf{MJ,\ \ \ \ }%
\mathbf{C}\left( \mathbf{q},\dot{\mathbf{q}}\right) =\mathsf{J}^{T}\mathsf{CJ%
}  \label{MC}
\end{equation}%
where%
\begin{eqnarray}
\mathsf{M} &:&=\mathrm{diag}\,(\mathbf{M}_{1},\ldots ,\mathbf{M}_{n})
\label{Msys} \\
\mathsf{C}(\mathbf{q},\dot{\mathbf{q}},\mathsf{V}\left( \dot{\mathbf{q}}%
\right) ) &:&=-\mathsf{MAa}-\mathsf{b}^{T}\mathsf{M}.  \label{Csys}
\end{eqnarray}%
and 
\begin{equation}
\mathsf{b}\left( \mathsf{V}\right) :=\mathrm{diag}~(\mathbf{ad}_{\mathbf{V}%
_{1}},\ldots ,\mathbf{ad}_{\mathbf{V}_{n}}).  \label{b}
\end{equation}%
Therein, the (constant) $6\times 6$ inertia matrix of body $i$ expressed in
the body-frame is defined as%
\begin{equation}
\mathbf{M}_{i}=\left( 
\begin{array}{cc}
{%
\bm{\Theta}%
_{i}^{\mathsf{b}}} & {^{i}}\widetilde{\mathbf{c}}_{i}m_{i} \\ 
-{^{i}}\widetilde{\mathbf{c}}_{i}m_{i}\ \ \  & m_{i}\mathbf{I}%
\end{array}%
\right)  \label{eqn_body_fixed_mass_inertia_matrix}
\end{equation}%
where $m_{i}$ is the body mass, ${^{i}}\mathbf{c}_{i}$ is the position
vector to the COM of body $i$ measured in the reference frame at body $i$,
and ${\bm{\Theta}_{i}^{\mathsf{b}}}$ is the inertia tensor with respect to
the body-fixed frame. The latter is related to the inertia tensor with
respect to the COM, denoted ${\bm{\Theta}_{i}^{\mathsf{c}}}$, via the
Steiner's theorem $\bm{\Theta}_{i}^{\mathsf{b}}=\mathbf{R}_{\mathsf{b},%
\mathsf{c}}\bm{\Theta}_{i}^{\mathsf{c}}\mathbf{R}_{\mathsf{b},\mathsf{c}%
}^{T}-m_{i}{^{i}}\widetilde{\mathbf{c}}_{i}{^{i}}\widetilde{\mathbf{c}}_{i}$
where $\mathbf{R}_{\mathsf{b},\mathsf{c}}$ denotes the rotation matrix of
the COM frame to the body frame.

A closed form of the EOM is obtained after replacing the system twist by (%
\ref{Vsys}). Alternatively, first the kinematic relation (\ref{Vsys}) and
then the coefficient matrices in (\ref{Csys}) are evaluated for a given
state $\mathbf{q},\dot{\mathbf{q}}$. The generalized gravity forces are
given as%
\begin{equation}
\mathbf{Q}_{\mathrm{grav}}\left( \mathbf{q}\right) =\mathsf{J}^{T}\mathsf{MU}%
\mathbf{G}
\end{equation}%
with%
\begin{equation}
\mathbf{G}:=-\left( 
\begin{array}{c}
\mathbf{0} \\ 
^{0}\mathbf{g}%
\end{array}%
\right) ,\ \ \mathsf{U}\left( \mathbf{q}\right) :=\mathsf{A}\left( 
\begin{array}{c}
\mathbf{Ad}_{\mathbf{C}_{1}}^{-1} \\ 
\mathbf{0} \\ 
\vdots \\ 
\mathbf{0}%
\end{array}%
\right) =\left( 
\begin{array}{c}
\mathbf{Ad}_{\mathbf{C}_{1}}^{-1} \\ 
\mathbf{Ad}_{\mathbf{C}_{2}}^{-1} \\ 
\vdots \\ 
\mathbf{Ad}_{\mathbf{C}_{n}}^{-1}%
\end{array}%
\right) .  \label{eq_U}
\end{equation}%
Here, $^{0}\mathbf{g}$ is the vector of gravitational acceleration expressed
in the inertial frame, which is transformed to the individual bodies by $%
\mathsf{U}$. The effect of contact or external wrenches acting on the bodies
is given by the generalized forces%
\begin{equation}
\mathbf{Q}_{\mathrm{ext}}\left( \mathbf{q},t\right) =\mathsf{J}^{T}\left( 
\mathbf{q}\right) \mathsf{W}_{\mathrm{EE}}\left( t\right) \quad \text{where}%
\quad \mathsf{W}_{\mathrm{ext}}\left( t\right) =\left( 
\begin{array}{c}
\mathbf{W}_{\mathrm{ext},1}\left( t\right) \\ 
\mathbf{W}_{\mathrm{ext},2}\left( t\right) \\ 
\vdots \\ 
\mathbf{W}_{\mathrm{ext},n}\left( t\right)%
\end{array}%
\right) .  \label{Qext}
\end{equation}%
The vector $\mathsf{W}_{\mathrm{ext},i}$ accounts for applied load at body $%
i $. For instance, $\mathbf{W}_{\mathrm{ext},n}$ can be used to describe the
wrench at the end-effector of a robotic arm. More generally, $\mathsf{W}_{%
\mathrm{ext}}$ and thus $\mathbf{Q}_{\mathrm{ext}}$, may account for
arbitrary (time and velocity dependent) loads at the system.

\section{First Time Derivative of the Equations of Motion}

\label{sec_first_order} The first time derivative of the generalized forces
is

\begin{equation}
\dot{\mathbf{Q}}=\mathbf{M}\dddot{\mathbf{q}}+(\dot{\mathbf{M}}+\mathbf{C})%
\ddot{\mathbf{q}}+\dot{\mathbf{C}}\dot{\mathbf{q}}+\dot{\mathbf{Q}}_{\mathrm{%
grav}}+\dot{\mathbf{Q}}_{\mathrm{ext}}.  \label{EOM1stOrder}
\end{equation}%
The time derivative of the generalized mass matrix $\mathbf{M}$ in (\ref{MC}%
) follows with $\dot{\mathsf{J}}$ in (\ref{Jdot}) as%
\begin{eqnarray}
\dot{\mathbf{M}}\left( \mathbf{q},\dot{\mathbf{q}}\right) &=&\mathsf{J}^{T}%
\mathsf{M}\dot{\mathsf{J}}+\dot{\mathsf{J}}^{T}\mathsf{MJ}  \notag \\
&=&\mathsf{J}^{T}\mathsf{M}^{\left( 1\right) }\mathsf{J}  \label{Mdot}
\end{eqnarray}%
with 
\begin{equation}
\mathsf{M}^{\left( 1\right) }\left( \mathbf{q},\dot{\mathbf{q}}\right) :=-%
\mathsf{MAa}-\left( \mathsf{MAa}\right) ^{T}  \label{M1}
\end{equation}%
where $\mathsf{A}=\mathsf{A}\left( \mathbf{q}\right) $ and $\mathsf{a}=%
\mathsf{a}\left( \dot{\mathbf{q}}\right) $. By the same token, the time
derivative of the generalized Coriolis matrix $\mathbf{C}$ is%
\begin{eqnarray}
\dot{\mathbf{C}}\left( \mathbf{q},\dot{\mathbf{q}},\ddot{\mathbf{q}}\right)
&=&\mathsf{J}^{T}\mathsf{C}\dot{\mathsf{J}}+\dot{\mathsf{J}}^{T}\mathsf{CJ}+%
\mathsf{J}^{T}\dot{\mathsf{C}}\mathsf{J}  \notag \\
&=&\mathsf{J}^{T}\mathsf{C}^{\left( 1\right) }\mathsf{J}  \label{Cdot}
\end{eqnarray}%
where now%
\begin{equation}
\mathsf{C}^{\left( 1\right) }:=\dot{\mathsf{C}}-\mathsf{CAa}-\mathsf{a}^{T}%
\mathsf{A}^{T}\mathsf{C}.  \label{C1}
\end{equation}%
The expression for $\mathsf{C}$ in (\ref{Csys}) together with (\ref{Jdot})
yields 
\begin{equation}
\dot{\mathsf{C}}(\mathbf{q},\dot{\mathbf{q}},\ddot{\mathbf{q}},\mathsf{\dot{%
\mathsf{V}}})=\mathsf{C}(\mathbf{q},\ddot{\mathbf{q}},\mathsf{\dot{\mathsf{V}%
}})+\mathsf{MAaAa}-\mathsf{MAaa}=\mathsf{MAaAa}-\mathsf{MAaa}-\mathsf{MA\dot{%
\mathsf{a}}-\dot{\mathsf{b}}}^{T}\mathsf{M},  \label{CSysdot}
\end{equation}%
with $\dot{\mathsf{a}}=\dot{\mathsf{a}}\left( \ddot{\mathbf{q}}\right) ,\dot{%
\mathsf{b}}=\mathsf{\dot{\mathsf{b}}}(\mathsf{\dot{\mathsf{V}}})$, and thus%
\begin{eqnarray}
\mathsf{C}^{\left( 1\right) } &=&\mathsf{C}(\mathbf{q},\ddot{\mathbf{q}},%
\mathsf{\dot{\mathsf{V}}})+\mathsf{MAaAa}-\mathsf{MAaa}-\mathsf{CAa}-\mathsf{%
a}^{T}\mathsf{A}^{T}\mathsf{C}  \label{C11} \\
&=&-\mathsf{MA\dot{\mathsf{a}}-\dot{\mathsf{b}}}^{T}\mathsf{M}+\mathsf{b}^{T}%
\mathsf{MAa}+\mathsf{a}^{T}\mathsf{A}^{T}\mathsf{b}^{T}\mathsf{M}+\mathsf{a}%
^{T}\mathsf{A}^{T}\mathsf{MAa}+2\mathsf{MAaAa}-  \notag \\
&&\mathsf{MAaa}.  \label{C12}
\end{eqnarray}%
The form (\ref{C11}) allows reusing the expression (\ref{Csys}) for the
matrix $\mathsf{C}$, where (\ref{Csys}) is evaluated with $\ddot{\mathbf{q}},%
\mathsf{\dot{\mathsf{V}}}$ instead with the velocities. Also in the second
form (\ref{C12}), terms like $\mathsf{Aa}$, $\mathsf{MAa}$, and $\mathsf{b}%
^{T}\mathsf{M}$ can be reused. A direct calculation yields%
\begin{equation}
\dot{\mathbf{Q}}_{\mathrm{grav}}\left( \mathbf{q},\dot{\mathbf{q}}\right) =%
\mathsf{J}^{T}\mathsf{M}^{\left( 1\right) }\mathsf{U}\mathbf{G}
\label{Qgravdot}
\end{equation}%
where the relation $\dot{\mathsf{U}}=-\mathsf{AaU}$ (obtained by time
differentiation of (\ref{eq_U}) and using the identity in (\ref{Adot})) and
the fact that $\mathbf{G}$ is constant has been used. The time derivative of
(\ref{Qext}) along with (\ref{Jdot}) yields the expression for calculating
first-order derivative of generalized forces due to external wrenches%
\begin{equation}
\dot{\mathbf{Q}}_{\mathrm{ext}}\left( \mathbf{q},\dot{\mathbf{q}}\right) =%
\mathsf{J}^{T}(\dot{\mathsf{W}}_{\mathrm{EE}}-(\mathsf{A}\mathsf{a})^{T}%
\mathsf{W}_{\mathrm{EE}}).  \label{Qextdot}
\end{equation}

\section{Second Time Derivative of the Equations of Motion}

\label{sec_second_order}

The second time derivative of the generalized motor forces (\ref{EOM1}) is
determined by%
\begin{equation}
\ddot{\mathbf{Q}}=\mathbf{M}\ddot{\ddot{\mathbf{q}}}+(2\dot{\mathbf{M}}+%
\mathbf{C})\dddot{\mathbf{q}}+(\ddot{\mathbf{M}}+2\dot{\mathbf{C}})\ddot{%
\mathbf{q}}+\ddot{\mathbf{C}}\dot{\mathbf{q}}+\ddot{\mathbf{Q}}_{\text{grav}%
}+\ddot{\mathbf{Q}}_{\mathrm{ext}}.  \label{EOM2ndOrder}
\end{equation}%
The second time derivative of the generalized mass matrix is%
\begin{eqnarray}
\ddot{\mathbf{M}}\left( \mathbf{q},\dot{\mathbf{q}},\ddot{\mathbf{q}}\right)
&=&\mathsf{J}^{T}(\dot{\mathsf{M}}^{\left( 1\right) }-\mathsf{M}^{\left(
1\right) }\mathsf{Aa}-\mathsf{a}^{T}\mathsf{A}^{T}\mathsf{M}^{\left(
1\right) })\mathsf{J} \\
&=&\mathsf{J}^{T}\mathsf{M}^{\left( 2\right) }\mathsf{J}  \label{Mddot}
\end{eqnarray}%
with $\mathsf{M}^{\left( 2\right) }:=\dot{\mathsf{M}}^{\left( 1\right) }-%
\mathsf{M}^{\left( 1\right) }\mathsf{Aa}-\mathsf{a}^{T}\mathsf{A}^{T}\mathsf{%
M}^{\left( 1\right) }$. Inserting the relation (\ref{Adot}) in the
differentiation of (\ref{M1}) yields%
\begin{eqnarray}
\dot{\mathsf{M}}^{\left( 1\right) }\left( \mathbf{q},\dot{\mathbf{q}},\ddot{%
\mathbf{q}}\right)  &=&\mathsf{M}^{\left( 1\right) }\left( \mathbf{q},\ddot{%
\mathbf{q}}\right) +\mathsf{M(AaA-Aa)a}+\left( \mathsf{M(AaA-Aa)a}\right)
^{T} \\
&=&\mathsf{M}^{\left( 1\right) }\left( \mathbf{q},\ddot{\mathbf{q}}\right) -%
\mathsf{MAaa}+\mathsf{MAaAa}-\left( \mathsf{MAaa}\right) ^{T}+\left( \mathsf{%
MAaAa}\right) ^{T}
\end{eqnarray}%
which leads immediately to the explicit form%
\begin{eqnarray}
\mathsf{M}^{\left( 2\right) } &=&\mathsf{M}^{\left( 1\right) }\left( \mathbf{%
q},\dot{\mathbf{q}}\right) +\mathsf{M(AaA-Aa)a}+\left( \mathsf{M(AaA-Aa)a}%
\right) ^{T} \notag \\
&&-\mathsf{M}^{\left( 1\right)}\left( \mathbf{%
q},\dot{\mathbf{q}}\right)\mathsf{\mathsf{Aa}}-(\mathsf{M}%
^{\left( 1\right) }\left( \mathbf{%
q},\dot{\mathbf{q}}\right)\mathsf{\mathsf{Aa}})^{T}  \label{M21} \\
&=&-\mathsf{MA\mathsf{\dot{\mathsf{a}}}}-\left( \mathsf{MA\mathsf{\dot{%
\mathsf{a}}}}\right) ^{T}+2\mathsf{\mathsf{MAaAa}}+2\left( \mathsf{\mathsf{%
\mathsf{MAaAa}}}\right) ^{T}+2\mathsf{\mathsf{a}}^{T}\mathsf{\mathsf{A}}^{T}%
\mathsf{\mathsf{MAa}} \\
&&-\mathsf{MAaa}-\left( \mathsf{MAaa}\right) ^{T}.  \notag  \label{M22}
\end{eqnarray}%
The expression (\ref{M21}) allows reusing $\mathsf{M}^{\left( 1\right)
}\left( \mathbf{q},\dot{\mathbf{q}}\right) $. It should be observed that the
term $\mathsf{M}^{\left( 1\right) }\left( \mathbf{q},\ddot{\mathbf{q}}%
\right) $ is the relation (\ref{M1}) evaluated with $\ddot{\mathbf{q}}$
instead of $\dot{\mathbf{q}}$.

Repeated time derivative of the generalized Coriolis matrix yields%
\begin{equation}
\ddot{\mathbf{C}}\left( \mathbf{q},\dot{\mathbf{q}},\ddot{\mathbf{q}},\dddot{%
\mathbf{q}}\right) =\mathsf{J}^{T}\mathsf{C}^{\left( 2\right) }\mathsf{J}
\end{equation}%
with%
\begin{equation}
\mathsf{C}^{\left( 2\right) }\left( \mathbf{q},\dot{\mathbf{q}},\ddot{%
\mathbf{q}},\dddot{\mathbf{q}}\right) =\dot{\mathsf{C}}^{\left( 1\right) }-%
\mathsf{C}^{\left( 1\right) }\mathsf{Aa}-\mathsf{a}^{T}\mathsf{A}^{T}\mathsf{%
C}^{\left( 1\right) }  \label{C2temp}
\end{equation}%
Taking the derivative of (\ref{C1}) yields $\dot{\mathsf{C}}^{\left(
1\right) }(\mathbf{q},\dot{\mathbf{q}},\ddot{\mathbf{q}},\dddot{\mathbf{q}},%
\mathsf{\dot{\mathsf{V}}},\ddot{\mathsf{V}})$ in closed form as%
\begin{eqnarray}
\dot{\mathsf{C}}^{\left( 1\right) } &=&\ddot{\mathsf{C}}-\dot{\mathsf{C}}%
\mathsf{Aa}-\mathsf{a}^{T}\mathsf{A}^{T}\dot{\mathsf{C}}-\mathsf{CA}\dot{%
\mathsf{a}}-\dot{\mathsf{a}}^{T}\mathsf{A}^{T}\mathsf{C}+\mathsf{CAaAa}+%
\mathsf{a}^{T}\mathsf{A}^{T}\mathsf{a}^{T}\mathsf{A}^{T}\mathsf{C}  \notag \\
&&-\mathsf{CAaa}-\mathsf{a}^{T}\mathsf{a}^{T}\mathsf{A}^{T}\mathsf{C}  \notag
\\
&=&\ddot{\mathsf{C}}-(\mathsf{C}^{\left( 1\right) }+\mathsf{a}^{T}\mathsf{A}%
^{T}\mathsf{C})\mathsf{Aa}-\mathsf{a}^{T}\mathsf{A}^{T}(\mathsf{C}^{\left(
1\right) }+\mathsf{CAa})-\mathsf{CA}\dot{\mathsf{a}}-\dot{\mathsf{a}}^{T}%
\mathsf{A}^{T}\mathsf{C} \\
&&-\mathsf{CAaa}-\mathsf{a}^{T}\mathsf{a}^{T}\mathsf{A}^{T}\mathsf{C}.\ \ \
\ \ \ \   \notag  \label{C1dot}
\end{eqnarray}%
This requires the derivative of (\ref{CSysdot}), which are found as%
\begin{eqnarray}
\ddot{\mathsf{C}} &=&\mathsf{C}(\mathbf{q},\dddot{\mathbf{q}},\mathsf{\ddot{%
\mathsf{V}}})+\mathsf{MA\dot{\mathsf{a}}Aa}+2\mathsf{MAaA\dot{\mathsf{a}}%
-2MAaAaAa}+\mathsf{2MAaAaa}-\mathsf{3MAa}\dot{\mathsf{a}} \\
&&-\mathsf{MAaaa}+\mathsf{MAaaAa}  \notag  \label{CSys2dot} \\
&=&-\mathsf{MA\ddot{\mathsf{a}}-\ddot{\mathsf{b}}}^{T}\mathsf{M}+\mathsf{MA%
\dot{\mathsf{a}}Aa}+2\mathsf{MAaA\dot{\mathsf{a}}-2MAaAaAa}+\mathsf{2MAaAaa}
\\
&&-\mathsf{3MAa}\dot{\mathsf{a}}-\mathsf{MAaaa}+\mathsf{MAaaAa}.  \notag
\end{eqnarray}%
Also here it should be observed that the matrix $\mathsf{C}(\mathbf{q},%
\dddot{\mathbf{q}},\mathsf{\ddot{\mathsf{V}}})$ is the expression for $%
\mathsf{C}$ in (\ref{Csys}) evaluated with $\dddot{\mathbf{q}},\mathsf{\ddot{%
\mathsf{V}}}$. Inserting (\ref{C1dot}) along with (\ref{CSys2dot}) into (\ref%
{C2temp}) yields $\mathsf{C}^{\left( 2\right) }\left( \mathbf{q},\dot{%
\mathbf{q}},\ddot{\mathbf{q}},\dddot{\mathbf{q}}\right) $ in the explicit
form%
\begin{eqnarray}
\mathsf{C}^{\left( 2\right) } &=&\mathsf{C}(\mathbf{q},\dddot{\mathbf{q}},%
\mathsf{\ddot{\mathsf{V}}})-2\mathsf{C}^{\left( 1\right) }\mathsf{Aa}-2%
\mathsf{a}^{T}\mathsf{A}^{T}\mathsf{C}^{\left( 1\right) }-\mathsf{CA}\dot{%
\mathsf{a}}-\dot{\mathsf{a}}^{T}\mathsf{A}^{T}\mathsf{C}-2\mathsf{a}^{T}%
\mathsf{A}^{T}\mathsf{CAa}  \notag \\
&&-\mathsf{CAaa}-\mathsf{a}^{T}\mathsf{a}^{T}\mathsf{A}^{T}\mathsf{C}+%
\mathsf{MA\dot{\mathsf{a}}Aa}+2\mathsf{MAaA}\dot{\mathsf{a}}-2\mathsf{MAaAaAa%
} \\
&&+\mathsf{2MAaAaa}-\mathsf{3MAa}\dot{\mathsf{a}}-\mathsf{MAaaa}+\mathsf{%
MAaaAa}  \notag  \label{C21} \\
&=&-\mathsf{MA}\ddot{\mathsf{a}}-\ddot{\mathsf{b}}^{T}\mathsf{M}+2\dot{%
\mathsf{b}}^{T}\mathsf{MAa}+2\mathsf{a}^{T}\mathsf{A}^{T}\dot{\mathsf{b}}^{T}%
\mathsf{M}  \notag \\
&&+3\mathsf{MA}\dot{\mathsf{a}}\mathsf{Aa}+2\mathsf{MAaA}\dot{\mathsf{a}}+2%
\mathsf{a}^{T}\mathsf{A}^{T}\mathsf{MA}\dot{\mathsf{a}}-\mathsf{CA}\dot{%
\mathsf{a}}-\dot{\mathsf{a}}^{T}\mathsf{A}^{T}\mathsf{C}  \notag \\
&&+2\mathsf{CAaAa}+2\mathsf{a}^{T}\mathsf{A}^{T}\mathsf{CAa}+2\mathsf{a}^{T}%
\mathsf{A}^{T}\mathsf{a}^{T}\mathsf{A}^{T}\mathsf{C}-4\mathsf{MAaAaAa} 
\notag \\
&&-2\mathsf{a}^{T}\mathsf{A}^{T}\mathsf{MAaAa}+\mathsf{2MAaaAa}+2\mathsf{a}%
^{T}\mathsf{A}^{T}\mathsf{MAaa} \\
&&+\mathsf{2MAaAaa}-\mathsf{3MAa}\dot{\mathsf{a}}-\mathsf{MAaaa}+\mathsf{%
MAaaAa}-\mathsf{CAaa}-\mathsf{a}^{T}\mathsf{a}^{T}\mathsf{A}^{T}\mathsf{C} 
\notag  \label{C22}
\end{eqnarray}%
with $\ddot{\mathsf{a}}=\ddot{\mathsf{a}}\left( \dddot{\mathbf{q}}\right) $
and $\ddot{\mathsf{b}}=\ddot{\mathsf{b}}(\ddot{\mathsf{V}})$ in (\ref{a})
and (\ref{b}) evaluated with $\dddot{\mathbf{q}}$ and $\ddot{\mathsf{V}}$,
respectively. The first form (\ref{C21}) may be beneficial since it involves 
$\mathsf{C}$ and $\mathsf{C}^{\left( 1\right) }$, which are already
available from (\ref{Csys}) and (\ref{C11}).

Taking the derivative of (\ref{Qgravdot}) yields the second time derivative
of the generalized gravity forces 
\begin{equation}
\ddot{\mathbf{Q}}_{\text{grav}}(\mathbf{q},\dot{\mathbf{q}},\ddot{\mathbf{q}}%
)=\mathsf{J}^{T}\mathsf{M}^{\left( 2\right) }\mathsf{U}\mathbf{G}.
\end{equation}%
The second time derivative of the generalized forces (\ref{Qext}) due to
end-effector loads is readily found as%
\begin{eqnarray}
\ddot{\mathbf{Q}}_{\mathrm{ext}}\left( \mathbf{q},\dot{\mathbf{q}},\ddot{%
\mathbf{q}}\right) &=&\mathsf{J}^{T}(\ddot{\mathsf{W}}_{\mathrm{EE}}-2(%
\mathsf{A}\mathsf{a})^{T}\dot{\mathsf{W}}_{\mathrm{EE}}+(2(\mathsf{A}\mathsf{%
a}\mathsf{A}\mathsf{a)}^{T}-(\mathsf{A}\dot{\mathsf{a}})^{T} \\
&&-\left( \mathsf{Aaa}\right) ^{T})\mathsf{W}_{\mathrm{EE}}).  \notag
\label{Qextddot}
\end{eqnarray}

\section{Structural Properties of the EOM}

\label{sec_structural_properties} There are two important and well-known
structural properties of EOM:

\begin{enumerate}
\item The generalized mass matrix $\mathbf{M}$ is symmetric and positive
definite.

\item $\dot{\mathbf{q}}^{T}\left( \dot{\mathbf{M}}\left( \mathbf{q},\dot{%
\mathbf{q}}\right) -2\mathbf{C}\left( \mathbf{q},\dot{\mathbf{q}}\right)
\right) \dot{\mathbf{q}}\equiv 0$ for any $\dot{\mathbf{q}}$
\end{enumerate}

Positive definiteness and the symmetric properties of the generalized mass
matrix follows directly from its definition in (\ref{MC}). The symmetry
property indeed applies to all higher derivatives, which is also evident
from (\ref{Mdot}) and (\ref{Mddot}). An important property of the EOM (\ref%
{EOM1}), which is crucial for proving stability of passivity-based control
schemes, is that the Coriolis matrix can be formulated as%
\begin{equation}
\bar{\mathbf{C}}:=-\mathsf{J}^{T}(\mathsf{MAa}+\mathsf{b}^{T}\mathsf{M}-%
\mathsf{Mb})\mathsf{J}  \label{Cbar}
\end{equation}%
so that $\dot{\mathbf{M}}-2\bar{\mathbf{C}}$ is skew symmetric. To show this
property, notice that $\mathsf{bJ}\dot{\mathbf{q}}=\mathsf{bV}=\mathrm{diag}%
\left( \mathbf{ad}_{\mathbf{V}_{1}}\mathbf{V}_{1},\ldots ,\mathbf{ad}_{%
\mathbf{V}_{n}}\mathbf{V}_{n}\right) \equiv \mathbf{0}$, so that $\mathbf{C}%
\dot{\mathbf{q}}$ can be rewritten as%
\begin{equation}
\mathbf{C}\dot{\mathbf{q}}=-\mathsf{J}^{T}(\mathsf{MAa}+\mathsf{b}^{T}%
\mathsf{M})\mathsf{J}\dot{\mathbf{q}}=-\mathsf{J}^{T}(\mathsf{MAa}+\mathsf{b}%
^{T}\mathsf{M}-\mathsf{Mb})\mathsf{J}\dot{\mathbf{q}}=\bar{\mathbf{C}}\dot{%
\mathbf{q}}
\end{equation}%
In view of the time derivative (\ref{Mdot}) of $\mathbf{M}$, the so
defined Coriolis matrix $\bar{\mathbf{C}}$ satisfies the relation%
\begin{equation}
\bar{\mathbf{C}}+\bar{\mathbf{C}}^{T}=-\mathsf{J}^{T}(\mathsf{MAa}+\mathsf{a}%
^{T}\mathsf{A}^{T}\mathsf{M})\mathsf{J}=\dot{\mathbf{M}}.
\end{equation}%
With (\ref{Cbar}), it hence follows%
\begin{equation}
\dot{\mathbf{M}}-2\bar{\mathbf{C}}=\bar{\mathbf{C}}^{T}-\bar{\mathbf{C}}=-(%
\bar{\mathbf{C}}^{T}-\bar{\mathbf{C}})^{T}=-(\dot{\mathbf{M}}-2\bar{\mathbf{C%
}})^{T}.
\end{equation}%
It should be remarked that the Coriolis matrix is not unique, and the
property 2) holds true for the particular arrangement as in (\ref{Csys}). It
may not hold if the equations are arranged differently. The skew symmetry of 
$\dot{\mathbf{M}}-2\bar{\mathbf{C}}$ is indeed carried over to its
derivatives.

\section{Examples}

\label{sec_results} The higher-order closed-form inverse dynamics
formulation presented in Section~\ref{sec_first_order} and Section~\ref%
{sec_second_order} were implemented in MATLAB\footnote{%
The source code as well as robot data are openly available at %
\url{https://github.com/shivesh1210/2nd_order_closed_form_time_derivatives_eom}%
}. This MATLAB implementation can also be used for efficient symbolic code
generation in MATLAB and C languages. This section presents the application
of this work to the computation of higher-order inverse dynamics of two
robot manipulators and presents a discussion on its computational efficiency.

\subsection{Planar 2R robot}

\begin{figure}[tbp]
\centering
\includegraphics[width=\textwidth]{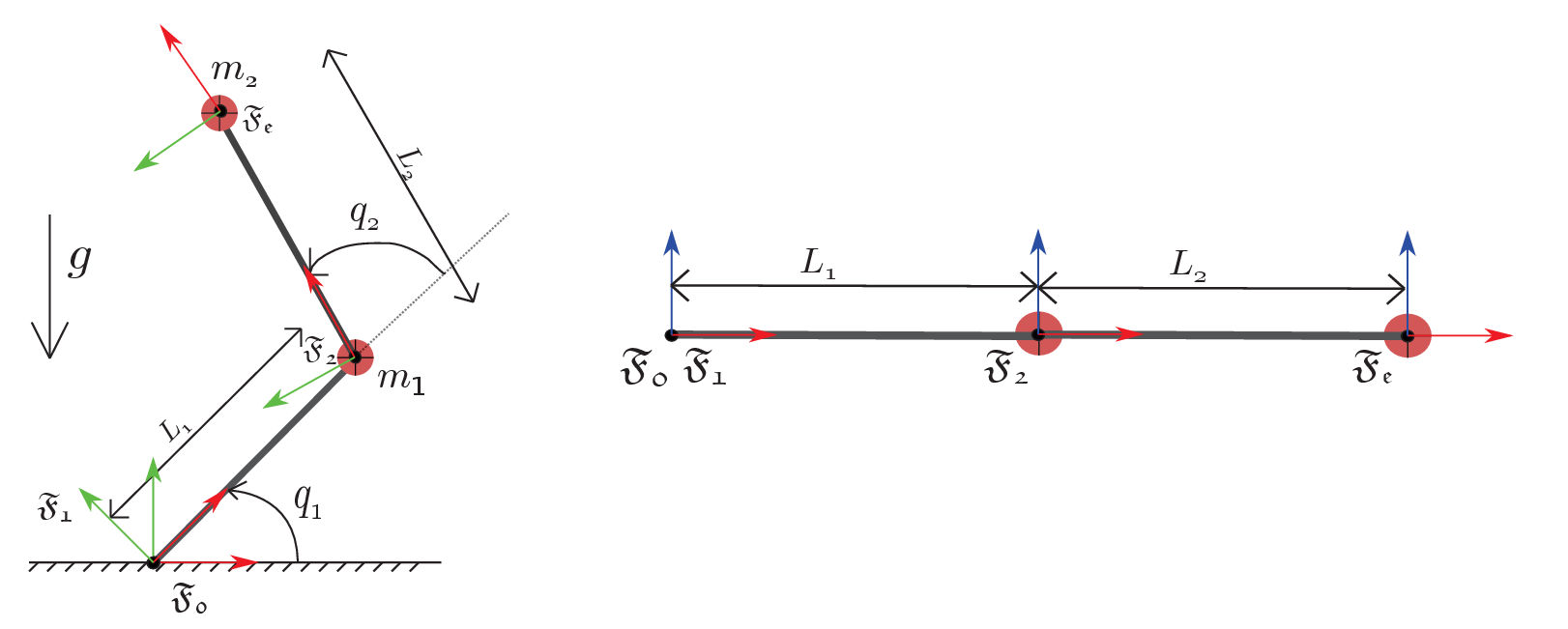}
\caption{Planar 2R robot schematic}
\label{fig_2r}
\end{figure}

A 2R serial chain consists of a base, two links, and two revolute joints. A
simple 2R chain is shown in Figure \ref{fig_2r}. First link of length $L_{1}$
is connected to the base or ground through a revolute joint. The second link
of length $L_{2}$ is connected to first link through a revolute joint. The
center of mass is shown with a red circle on the links and lies at the end
of each link. The mass of first and second links are $m_{1}$ and $m_{2}$
respectively. The gravity acting on the system is shown as $g$ in the figure.

\subsubsection{Kinematic model}

The $z$--axes represent the joint axes, and thus the joint screw coordinates
in body-fixed representation for the two revolute joints are%
\begin{eqnarray*}
{^{1}\mathbf{X}}_{1} &=&\left( 0,0,1,0,0,0\right) ^{T} \\
{^{2}\mathbf{X}}_{2} &=&\left( 0,0,1,0,0,0\right) ^{T}.
\end{eqnarray*}%
The relative reference configurations for the two bodies are%
\begin{equation*}
\mathbf{B}_{1}=\left( 
\begin{array}{cccc}
1 & 0 & 0 & 0 \\ 
0 & 1 & 0 & 0 \\ 
0 & 0 & 1 & 0 \\ 
0 & 0 & 0 & 1%
\end{array}%
\right) ,\quad \mathbf{B}_{2}=\left( 
\begin{array}{cccc}
1 & 0 & 0 & L_{1} \\ 
0 & 1 & 0 & 0 \\ 
0 & 0 & 1 & 0 \\ 
0 & 0 & 0 & 1%
\end{array}%
\right) .
\end{equation*}%
The mass matrices of the two bodies in the body fixed configuration
according to~(\ref{eqn_body_fixed_mass_inertia_matrix}) are%
\begin{eqnarray*}
\mathbf{M}_{1} &=&\left( 
\begin{array}{cccccc}
{L_{1}}^{2}\,m_{1} & 0 & 0 & 0 & 0 & 0 \\ 
0 & {L_{1}}^{2}\,m_{1} & 0 & 0 & 0 & -L_{1}\,m_{1} \\ 
0 & 0 & {L_{1}}^{2}\,m_{1} & 0 & L_{1}\,m_{1} & 0 \\ 
0 & 0 & 0 & m_{1} & 0 & 0 \\ 
0 & 0 & L_{1}\,m_{1} & 0 & m_{1} & 0 \\ 
0 & -L_{1}\,m_{1} & 0 & 0 & 0 & m_{1}%
\end{array}%
\right) , \\
\mathbf{M}_{2} &=&\left( 
\begin{array}{cccccc}
{L_{2}}^{2}\,m_{2} & 0 & 0 & 0 & 0 & 0 \\ 
0 & {L_{2}}^{2}\,m_{2} & 0 & 0 & 0 & -L_{2}\,m_{2} \\ 
0 & 0 & {L_{2}}^{2}\,m_{2} & 0 & L_{2}\,m_{2} & 0 \\ 
0 & 0 & 0 & m_{2} & 0 & 0 \\ 
0 & 0 & L_{2}\,m_{2} & 0 & m_{2} & 0 \\ 
0 & -L_{2}\,m_{2} & 0 & 0 & 0 & m_{2}%
\end{array}%
\right) .
\end{eqnarray*}

\subsubsection{Second-order inverse dynamics}

With the above information about the robot, using the closed form
expressions of $\mathbf{M}(\mathbf{q}),\mathbf{C}(\mathbf{q},\mathbf{\dot{q}}%
)$ and $\mathbf{Q}_{\text{grav}}(\mathbf{q})$ provided in Section~\ref%
{sec_eom_formulae} in (\ref{EOM1}), one can arrive at the analytical
expression for generalized forces $(\tau _{1},\tau _{2})$ which solves
inverse dynamics problem as follows%
\begin{eqnarray*}
\tau _{1} &=&{L_{1}}^{2}\,\ddot{q}_{1}\,m_{1}+{L_{1}}^{2}\,\ddot{q}%
_{1}\,m_{2}+{L_{2}}^{2}\,\ddot{q}_{1}\,m_{2}+{L_{2}}^{2}\,\ddot{q}%
_{2}\,m_{2}+L_{2}\,g\,m_{2}\,\cos \left( q_{1}+q_{2}\right) \\
&&+L_{1}\,g\,m_{1}\,\cos \left( q_{1}\right) +L_{1}\,g\,m_{2}\,\cos \left(
q_{1}\right) -L_{1}\,L_{2}\,{\dot{q}_{2}}^{2}\,m_{2}\,\sin \left(
q_{2}\right) \\
&&+2\,L_{1}\,L_{2}\,\ddot{q}_{1}\,m_{2}\,\cos \left( q_{2}\right)
+L_{1}\,L_{2}\,\ddot{q}_{2}\,m_{2}\,\cos \left( q_{2}\right)
-2\,L_{1}\,L_{2}\,\dot{q}_{1}\,\dot{q}_{2}\,m_{2}\,\sin \left( q_{2}\right)
\\
\tau _{2} &=&L_{2}\,m_{2}\,\left( L_{1}\,\sin \left( q_{2}\right) \,{\dot{q}%
_{1}}^{2}+L_{2}\,\ddot{q}_{1}+L_{2}\,\ddot{q}_{2}+g\,\cos \left(
q_{1}+q_{2}\right) +L_{1}\,\ddot{q}_{1}\,\cos \left( q_{2}\right) \right) .
\end{eqnarray*}%
These expressions are indeed identical to those reported in~\cite%
{LynchPark2017} for instance. 

Using the closed form expressions of $\dot{\mathbf{M}}(\mathbf{q,\dot{q}}),%
\dot{\mathbf{C}}(\mathbf{q},\mathbf{\dot{q}},\mathbf{\ddot{q}})$ and $\dot{%
\mathbf{Q}}_{\text{grav}}(\mathbf{q},\mathbf{\dot{q}})$ provided in Section~%
\ref{sec_first_order} in (\ref{EOM1stOrder}), one obtains the analytical
expression for first-order time derivatives $(\dot{\tau}_{1},\dot{\tau}_{2})$
of the generalized forces, which solves the first-order inverse dynamics
problem, as follows%
\begin{eqnarray*}
\dot{\tau}_{1} &=&{L_{1}}^{2}\,\dddot{q}_{1}\,m_{1}+{L_{1}}^{2}\,\dddot{q}%
_{1}\,m_{2}+{L_{2}}^{2}\,\dddot{q}_{1}\,m_{2}+{L_{2}}^{2}\,\dddot{q}%
_{2}\,m_{2}-L_{2}\,\dot{q}_{1}\,g\,m_{2}\,\sin \left( q_{1}+q_{2}\right) \\
&&-L_{2}\,\dot{q}_{2}\,g\,m_{2}\,\sin \left( q_{1}+q_{2}\right)
+2\,L_{1}\,L_{2}\,\dddot{q}_{1}\,m_{2}\,\cos \left( q_{2}\right)
+L_{1}\,L_{2}\,\dddot{q}_{2}\,m_{2}\,\cos \left( q_{2}\right) \\
&&-L_{1}\,\dot{q}_{1}\,g\,m_{1}\,\sin \left( q_{1}\right) -L_{1}\,\dot{q}%
_{1}\,g\,m_{2}\,\sin \left( q_{1}\right) -L_{1}\,L_{2}\,{\dot{q}_{2}}%
^{3}\,m_{2}\,\cos \left( q_{2}\right) \\
&&-4\,L_{1}\,L_{2}\,\ddot{q}_{1}\,\dot{q}_{2}\,m_{2}\,\sin \left(
q_{2}\right) -2\,L_{1}\,L_{2}\,\ddot{q}_{2}\,\dot{q}_{1}\,m_{2}\,\sin \left(
q_{2}\right) \\
&&-3\,L_{1}\,L_{2}\,\ddot{q}_{2}\,\dot{q}_{2}\,m_{2}\,\sin \left(
q_{2}\right) -2\,L_{1}\,L_{2}\,\dot{q}_{1}\,{\dot{q}_{2}}^{2}\,m_{2}\,\cos
\left( q_{2}\right) \\
\dot{\tau}_{2} &=&L_{2}\,m_{2}\,(L_{2}\,\dddot{q}_{1}+L_{2}\,\dddot{q}_{2}-%
\dot{q}_{1}\,g\,\sin \left( q_{1}+q_{2}\right) -\dot{q}_{2}\,g\,\sin \left(
q_{1}+q_{2}\right) \\
&&+L_{1}\,\dddot{q}_{1}\,\cos \left( q_{2}\right) +2\,L_{1}\,\ddot{q}_{1}\,%
\dot{q}_{1}\,\sin \left( q_{2}\right) -L_{1}\,\ddot{q}_{1}\,\dot{q}%
_{2}\,\sin \left( q_{2}\right) +L_{1}\,{\dot{q}_{1}}^{2}\,\dot{q}_{2}\,\cos
\left( q_{2}\right) ).
\end{eqnarray*}%
The correctness can be easily checked taking the analytical time derivative
of $(\tau _{1},\tau _{2})$ above.

Using the closed form expressions of $\ddot{\mathbf{M}}(\mathbf{q,\dot{q},%
\ddot{q}}),\ddot{\mathbf{C}}(\mathbf{q},\mathbf{\dot{q}},\mathbf{\ddot{q}},%
\mathbf{\dddot{q}})$ and $\ddot{\mathbf{Q}}_{\text{grav}}(\mathbf{q},\mathbf{%
\dot{q}},\mathbf{\ddot{q}})$ provided in Section~\ref{sec_second_order} in (%
\ref{EOM2ndOrder}), one obtains the analytical expression for second-order
time derivatives of generalized forces $(\ddot{\tau}_{1},\ddot{\tau}_{2})$,
solving the second-order inverse dynamics problem: 
\begin{eqnarray*}
\ddot{\tau}_{1} &=&{L_{1}}^{2}\,\ddot{\ddot{q}}_{1}\,m_{1}+{L_{1}}^{2}\,%
\ddot{\ddot{q}}_{1}\,m_{2}+{L_{2}}^{2}\,\ddot{\ddot{q}}_{1}\,m_{2}+{L_{2}}%
^{2}\,\ddot{\ddot{q}}_{2}\,m_{2}-3\,L_{1}\,L_{2}\,{\ddot{q}_{2}}%
^{2}\,m_{2}\,\sin \left( q_{2}\right) \\
&&+L_{1}\,L_{2}\,{\dot{q}_{2}}^{4}\,m_{2}\,\sin \left( q_{2}\right) -L_{1}\,{%
\dot{q}_{1}}^{2}\,g\,m_{1}\,\cos \left( q_{1}\right) -L_{1}\,{\dot{q}_{1}}%
^{2}\,g\,m_{2}\,\cos \left( q_{1}\right) \\
&&-L_{2}\,\ddot{q}_{1}\,g\,m_{2}\,\sin \left( q_{1}+q_{2}\right) -L_{2}\,%
\ddot{q}_{2}\,g\,m_{2}\,\sin \left( q_{1}+q_{2}\right) +2\,L_{1}\,L_{2}\,%
\ddot{\ddot{q}}_{1}\,m_{2}\,\cos \left( q_{2}\right) \\
&&+L_{1}\,L_{2}\,\ddot{\ddot{q}}_{2}\,m_{2}\,\cos \left( q_{2}\right)
-L_{1}\,\ddot{q}_{1}\,g\,m_{1}\,\sin \left( q_{1}\right) -L_{1}\,\ddot{q}%
_{1}\,g\,m_{2}\,\sin \left( q_{1}\right) \\
&&-L_{2}\,{\dot{q}_{1}}^{2}\,g\,m_{2}\,\cos \left( q_{1}+q_{2}\right)
-L_{2}\,{\dot{q}_{2}}^{2}\,g\,m_{2}\,\cos \left( q_{1}+q_{2}\right)
-6\,L_{1}\,L_{2}\,\ddot{q}_{1}\,\ddot{q}_{2}\,m_{2}\,\sin \left( q_{2}\right)
\\
&&-6\,L_{1}\,L_{2}\,\dddot{q}_{1}\,\dot{q}_{2}\,m_{2}\,\sin \left(
q_{2}\right) -2\,L_{1}\,L_{2}\,\dddot{q}_{2}\,\dot{q}_{1}\,m_{2}\,\sin
\left( q_{2}\right) -4\,L_{1}\,L_{2}\,\dddot{q}_{2}\,\dot{q}%
_{2}\,m_{2}\,\sin \left( q_{2}\right) \\
&&-6\,L_{1}\,L_{2}\,\ddot{q}_{1}\,{\dot{q}_{2}}^{2}\,m_{2}\,\cos \left(
q_{2}\right) -6\,L_{1}\,L_{2}\,\ddot{q}_{2}\,{\dot{q}_{2}}^{2}\,m_{2}\,\cos
\left( q_{2}\right) +2\,L_{1}\,L_{2}\,\dot{q}_{1}\,{\dot{q}_{2}}%
^{3}\,m_{2}\,\sin \left( q_{2}\right) \\
&&-2\,L_{2}\,\dot{q}_{1}\,\dot{q}_{2}\,g\,m_{2}\,\cos \left(
q_{1}+q_{2}\right) -6\,L_{1}\,L_{2}\,\ddot{q}_{2}\,\dot{q}_{1}\,\dot{q}%
_{2}\,m_{2}\,\cos \left( q_{2}\right) \\
\ddot{\tau}_{2} &=&-L_{2}\,m_{2}\,({\dot{q}_{1}}^{2}\,g\,\cos \left(
q_{1}+q_{2}\right) -L_{2}\,\ddot{\ddot{q}}_{2}-L_{2}\,\ddot{\ddot{q}}_{1}+{%
\dot{q}_{2}}^{2}\,g\,\cos \left( q_{1}+q_{2}\right) \\
&&-2\,L_{1}\,{\ddot{q}_{1}}^{2}\,\sin \left( q_{2}\right) +\ddot{q}%
_{1}\,g\,\sin \left( q_{1}+q_{2}\right) +\ddot{q}_{2}\,g\,\sin \left(
q_{1}+q_{2}\right) -L_{1}\,\ddot{\ddot{q}}_{1}\,\cos \left( q_{2}\right) \\
&&+L_{1}\,{\dot{q}_{1}}^{2}\,{\dot{q}_{2}}^{2}\,\sin \left( q_{2}\right) +2\,%
\dot{q}_{1}\,\dot{q}_{2}\,g\,\cos \left( q_{1}+q_{2}\right) +L_{1}\,\ddot{q}%
_{1}\,\ddot{q}_{2}\,\sin \left( q_{2}\right) -2\,L_{1}\,\dddot{q}_{1}\,\dot{q%
}_{1}\,\sin \left( q_{2}\right) \\
&&+2\,L_{1}\,\dddot{q}_{1}\,\dot{q}_{2}\,\sin \left( q_{2}\right) +L_{1}\,%
\ddot{q}_{1}\,{\dot{q}_{2}}^{2}\,\cos \left( q_{2}\right) -L_{1}\,\ddot{q}%
_{2}\,{\dot{q}_{1}}^{2}\,\cos \left( q_{2}\right) -4\,L_{1}\,\ddot{q}_{1}\,%
\dot{q}_{1}\,\dot{q}_{2}\,\cos \left( q_{2}\right) ).
\end{eqnarray*}%
They can again be verified by taking analytic derivatives of the above $(%
\dot{\tau}_{1},\dot{\tau}_{2})$.

\begin{figure}[tb]
\centering
\includegraphics[width=0.75\linewidth]{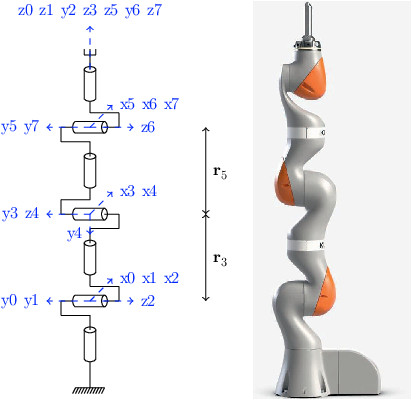}
\caption{Link frames of the KUKA LBR iiwa 14 R820 robot~\protect\cite{iiwa}}
\label{fig_iiwa_robot}
\end{figure}

\subsection{KUKA LBR iiwa manipulator}

The closed-form formulation presented in this paper are used to compute the
second-order inverse dynamics of the 7 degrees of freedom KUKA LBR iiwa
robot.

\subsubsection{Kinematic model}

The body-fixed frames are introduced according to~\cite{iiwa}, which follow
the modified Denavit-Hartenberg (DH) convention. As the $z$--axis represents
the joint axis in this convention, the joint screw coordinates in body-fixed
representation are%
\begin{equation}
{^{i}\mathbf{X}}_{i}=\left( 0,0,1,0,0,0\right) ^{T},i=1,\ldots ,7.
\end{equation}%
The relative reference configurations ${\mathbf{B}}_{i}$ in (\ref{Ci}) have
the form (\ref{Cij}), with the relative rotation matrix $\mathbf{R}_{i-1,i}$
and the position vector ${^{i-1}}\mathbf{r}_{i-1,i}$ from the origin of the
reference of body $i-1$ to that of its successor body $i$. According to the
zero reference configuration shown in Figure~\ref{fig_iiwa_robot} in \cite%
{iiwa}, the body-fixed displacement vectors are%
\begin{eqnarray*}
{^{i-1}}\mathbf{r}_{i-1,i} &=&\mathbf{0},\quad i=1,2,4,6,7 \\
{^{2}}\mathbf{r}_{2,3} &=&\left( 0,r_{3},0\right) ^{T},{^{4}}\mathbf{r}%
_{4,5}=\left( 0,-r_{5},0\right) ^{T}
\end{eqnarray*}%
and the relative rotation matrices are%
\begin{eqnarray*}
\mathbf{R}_{0,1} &=&\mathbf{I} \\
\mathbf{R}_{i-1,i} &=&\left( 
\begin{array}{ccc}
1 & 0 & 0 \\ 
0 & 0 & -1 \\ 
0 & 1 & 0%
\end{array}%
\right) ,\quad i=2,5,6 \\
\mathbf{R}_{i-1,i} &=&\left( 
\begin{array}{ccc}
1 & 0 & 0 \\ 
0 & 0 & 1 \\ 
0 & -1 & 0%
\end{array}%
\right) ,\quad i=3,4,7.
\end{eqnarray*}

\subsubsection{Dynamic model parameters}

The mass and inertia data used in this example is taken from the identified
model of KUKA LBR iiwa robot reported in~\cite{iiwa}. The COM position
vector $^{i}\mathbf{c}_{i}=({^{i}}c_{i,x},{^{i}}c_{i,y},{^{i}}%
c_{i,z})^{T}\in \mathbb{R}^{3}$ of link $i$ is measured with respect to the
link $i$ frame. The parameters in the symmetric inertia tensor of link $i$
are defined relative to the COM of link $i$ in the $i^{\text{th}}$ link
frame.

\subsubsection{Second-order inverse dynamics}

\begin{figure}[tbp]
\centering
\begin{subfigure}[b]{0.95\textwidth}
         \centering
         \includegraphics[width=\textwidth]{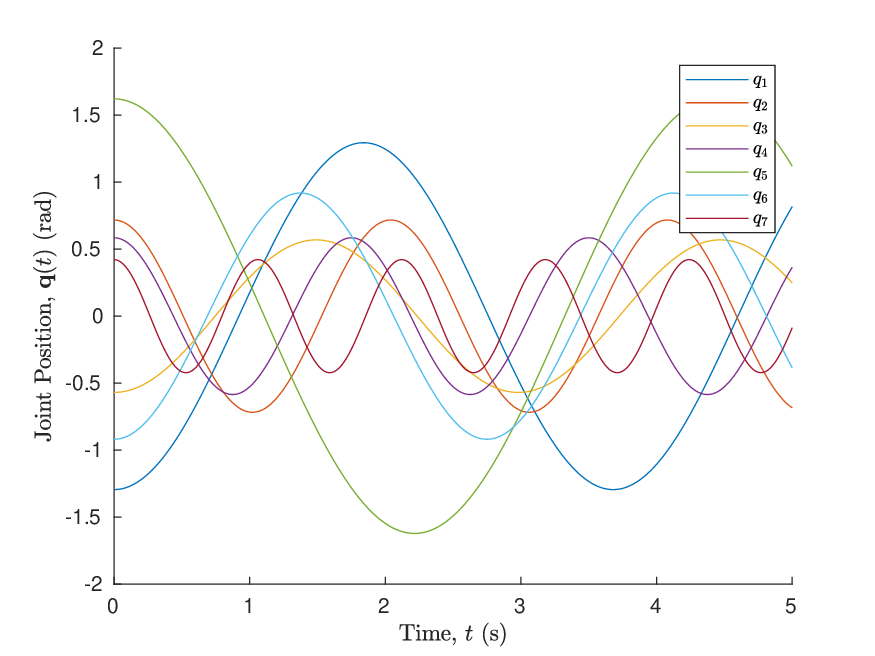}
         \caption{Joint motion trajectories $\mathbf{q}(t)$}
         \label{fig:input_motion}
     \end{subfigure}
\hfill 
\begin{subfigure}[b]{0.95\textwidth}
         \centering
         \includegraphics[width=\textwidth]{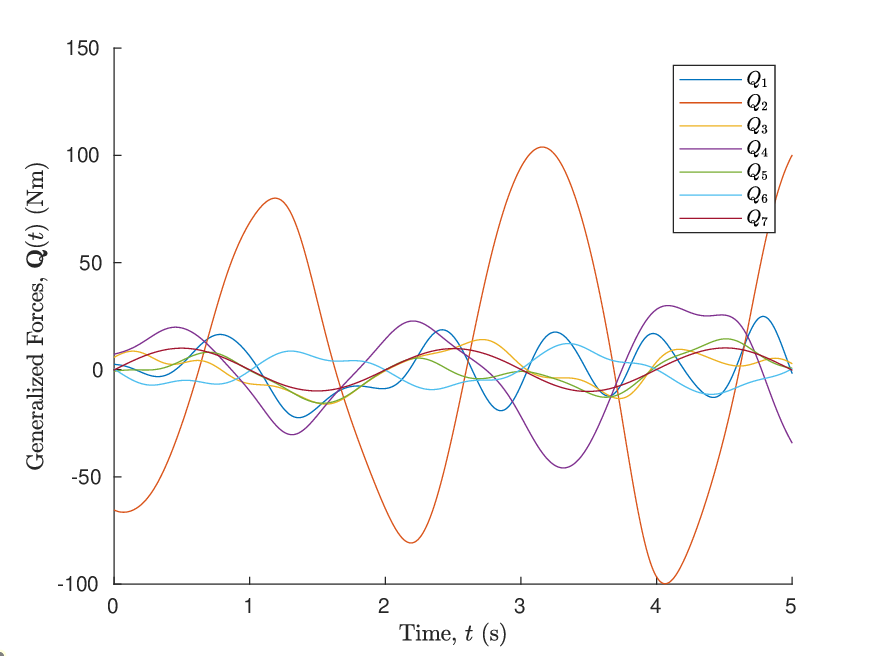}
         \caption{Inverse dynamics output $\mathbf{Q}(t)$}
         \label{fig:inv_dyn}
     \end{subfigure}
\caption{Joint trajectories and corresponding joint torques of the KUKA LBR
iiwa 14 R820 robot}
\label{fig_input_motion_and_inv_dynamics}
\end{figure}

\begin{figure}[tbp]
\centering
\begin{subfigure}[b]{0.95\textwidth}
         \centering
         \includegraphics[width=\textwidth]{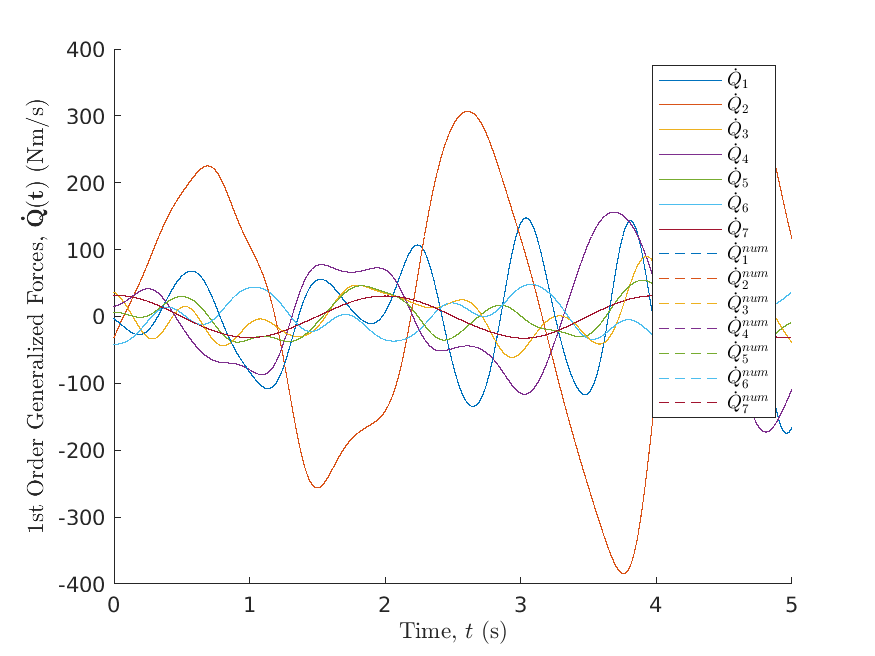}
         \caption{1st order inverse dynamics $\mathbf{\dot{Q}}(t)$}
     \end{subfigure}
\hfill 
\begin{subfigure}[b]{0.95\textwidth}
         \centering
         \includegraphics[width=\textwidth]{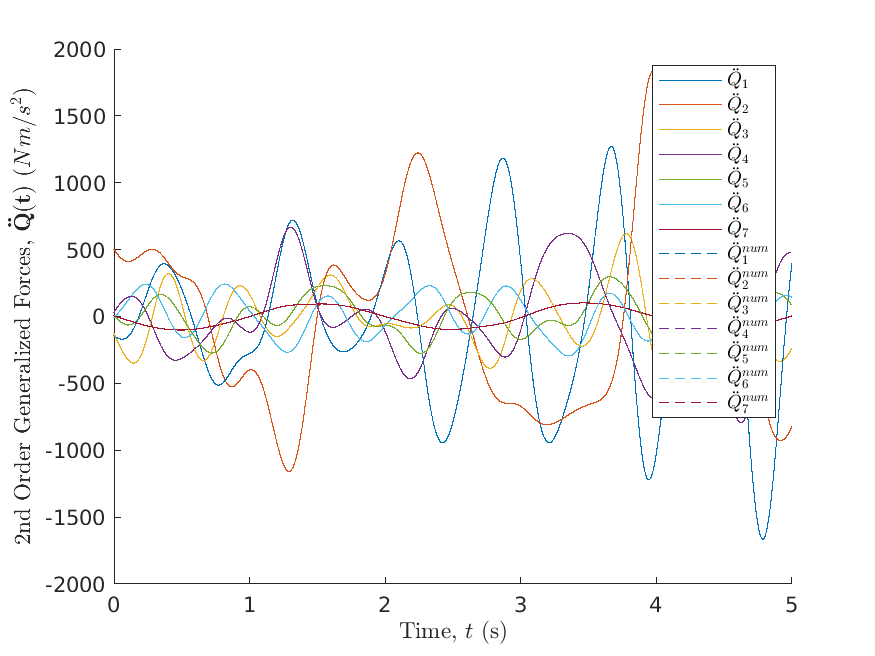}
         \caption{2nd order inverse dynamics $\mathbf{\ddot{Q}}(t)$}
     \end{subfigure}
\caption{Results for the higher-order inverse dynamics}
\label{fig_derivatives_inv_dynamics}
\end{figure}

The joint trajectory is described by means of cos-function as shown in
Figure~\ref{fig:input_motion}, and used for the inverse dynamics computation
of the robotic manipulator (see Figure~\ref{fig:inv_dyn}). Figure~\ref%
{fig:input_motion} and Figure~\ref{fig:inv_dyn} show first- and second-order
time derivatives of the generalized forces. The results computed from the
closed form expressions are compared against the numerical differentiation.
As apparent from Figure~\ref{fig_derivatives_inv_dynamics}, the closed form
derivatives match the numerical time differentiation of generalized forces
in both cases which attests the correctness of the formulation presented in
this paper.

\subsubsection{Computational Performance}

The computational performance of the closed form algorithm was evaluated by
measuring the total CPU time spent in 10000 evaluations of second-order
inverse dynamics on a standard laptop with Intel Core i9-7960X CPU clocked
at 2.80 GHz and 128 GB RAM. It was found that the closed form implementation
takes $16.34$ seconds leading to the average computational time of $1.6$
milliseconds per evaluation which is sufficient for many real--time control
applications.

\begin{table}[tb]
\centering 
\begin{tabular}{ccccccccccc}
\hline
$i$ & $m^{i}$ & $^{i}c_{i,x}$ & $^{i}c_{i,y}$ & $^{i}c_{i,z}$ & $\Theta
_{i,xx}^{\mathrm{c}}$ & $\Theta _{i,xy}^{\mathrm{c}}$ & $\Theta _{i,yx}^{%
\mathrm{c}}$ & $\Theta _{i,yy}^{\mathrm{c}}$ & $\Theta _{i,yz}^{\mathrm{c}}$
& $\Theta _{i,zz}^{\mathrm{c}}$ \\ \hline
1 & 3.94781 & -0.00351 & 0.00160 & -0.03139 & 0.00455 & 0.00000 & -0.00000 & 
0.00454 & 0.00001 & 0.00029 \\ 
2 & 4.50275 & -0.00767 & 0.16669 & -0.00355 & 0.00032 & 0.00000 & 0.00000 & 
0.00010 & -0.00000 & 0.00042 \\ 
3 & 2.45520 & -0.00225 & -0.03492 & -0.02652 & 0.00223 & -0.00005 & 0.00007
& 0.00219 & 0.00007 & 0.00073 \\ 
4 & 2.61155 & 0.00020 & -0.05268 & 0.03818 & 0.03844 & 0.00088 & -0.00112 & 
0.01144 & -0.00111 & 0.04988 \\ 
5 & 3.41000 & 0.00005 & -0.00237 & -0.21134 & 0.00277 & -0.00001 & 0.00001 & 
0.00284 & -0.00000 & 0.00012 \\ 
6 & 3.38795 & 0.00049 & 0.02019 & -0.02750 & 0.00050 & -0.00005 & -0.00003 & 
0.00281 & -0.00004 & 0.00232 \\ 
7 & 0.35432 & -0.03466 & -0.02324 & 0.07138 & 0.00795 & 0.00022 & -0.00029 & 
0.01089 & -0.00029 & 0.00294 \\ \hline
\end{tabular}%
\caption{Inertia data of different bodies in the KUKA robot as identified in~%
\protect\cite{iiwa}}
\end{table}

\section{Conclusion and Outlook}

\label{sec_conclusion} This paper presents closed form expressions for the
first and second time derivative of the EOM of a kinematic chain. Building
upon the Lie group formulation of the EOM, the formulations are advantageous
as they are expressed in terms of joint screw coordinates, and thus
facilitate parameterization in terms of vector quantities that can be easily
obtained. The computationally efficiency of these closed form relations
compared to recursive algorithms and their efficient implementation will be
a topic of further research. It is already obvious from the presented
expressions that they involve many repeated terms that can be precomputed
and reused. Future research will also address the time derivatives of
general mechanisms with kinematic loops and an efficient C++ based
implementation in Hybrid Robot Dynamics (HyRoDyn) software framework~\cite%
{KumarHyRoDyn2020}.

\subsubsection*{Acknowledgments}

This work has been supported by the LCM K2 Center for Symbiotic Mechatronics
within the framework of the Austrian COMET-K2 program. The second author
acknowledges the support of Q-RoCK (Grant Number: FKZ 01IW18003) and
VeryHuman (Grant Number: FKZ 01IW20004) projects funded by the German
Aerospace Center (DLR) with federal funds from the Federal Ministry of
Education and Research (BMBF).

The authors further thank Rohit Kumar from the University Genoa for helping with the implementation and testing of the symbolic code.

\end{document}